# Applying LLM and Topic Modelling in Psychotherapeutic Contexts

**Alexander Vanin [1], Vadim Bolshev [2,*], Anastasia Panfilova [3]**

[1] Laborarory of AI application in Psychology, Institute of Psychology (RAS), e-mail: vaninav@ipran.ru
[2] Laborarory of AI application in Psychology, Institute of Psychology (RAS), e-mail: bolshevv@ipran.ru
[3] Laborarory of AI application in Psychology, Institute of Psychology (RAS), e-mail: panfilova87@gmail.com

## Abstract

This study explores the use of LLM (Large language models) to analyze therapist remarks in a psychotherapeutic setting. The paper focuses on the application of BERTopic, a machine learning-based topic modeling tool, to the dialogue of two different groups of therapists—classical and modern—which makes it possible to identify and describe a set of topics that consistently emerge across these groups. The paper describes in detail the chosen algorithm for BERTopic, which included creating a vector space from a corpus of therapist remarks, reducing its dimensionality, clustering the space, and creating and optimizing topic representation. Along with the automatic topical modeling by the BERTopic, the research involved an expert assessment of the findings and manual topic structure optimization. The topic modeling results highlighted the most common and stable topics in therapists' speech, offering insights into how language patterns in therapy develop and remain stable across different therapeutic styles. This work contributes to the growing field of machine learning in psychotherapy by demonstrating the potential of automated methods to improve both the practice and training of therapists. The study highlights the value of topic modeling as a tool for gaining a deeper understanding of therapeutic dialogue and offers new opportunities for improving therapeutic effectiveness and clinical supervision.

## 1. Introduction

In psychotherapy, the language and methods therapists use play a crucial role in shaping the therapeutic process and influencing client outcomes. Analyzing how therapists communicate, through their interventions, strategies, and speech patterns, can provide valuable insights into therapy's effectiveness. Traditionally, this analysis relied on manual coding, a labor-intensive process prone to bias (Wampold & Imel, 2015). However, advancements in machine learning, particularly in topic modeling, have made it possible to analyze large volumes of textual data efficiently and without such biases.

This study explores the use of BERTopic, a machine learning technique that leverages transformer-based models to identify and extract meaningful topics from unstructured datasets. BERTopic offers a powerful tool for analyzing the complex themes that emerge during therapy sessions. Topic modeling methods, such as LDA (Blei et al., 2001) and Top2Vec (Angelov, 2020), have proven effective in large-scale textual analysis within social science research (Gaur et al., 2018; Yanchuk et al., 2022). Furthermore, BERT-based models have been applied in mental health research to predict outcomes and identify patterns in psychotherapeutic data (Zeberga et al., 2022). This study aims to assess the effectiveness of BERTopic (Grootendorst, 2022) in automatically extracting relevant topics from psychotherapeutic dialogues, providing deeper insights into the therapeutic process and therapists' strategies.

Topic modeling offers several potential benefits in psychotherapy. It can identify which therapeutic techniques – such as empathy, reframing, or cognitive restructuring – are most frequently used by therapists, thereby informing clinical practice, supervision, and training (Norcross & Wampold, 2011). Additionally, topic modeling can track shifts in therapeutic approaches over time, enabling data-driven evaluations of treatment progress and adjustments to better meet clients' evolving needs (Lambert & Ogles, 2004).

This article examines BERTopic's capabilities in analyzing therapist remarks, evaluating its effectiveness in identifying meaningful topics. Ultimately, the study aims to contribute to the growing body of research on machine learning applications in psychotherapy, providing new opportunities to enhance clinical practice and therapist training.

## 2. Related Work

Recent research in machine learning and psychotherapy has focused on analyzing therapeutic dialogues, often emphasizing patients' language or the overall therapist-client interaction. For example, Gao and Sazara (2023) (Gao & Sazara, 2023) employed a customized Sentence-BERT model integrated with the BERTopic framework to analyze over 96,000 research paper abstracts in mental health. Their findings reveal that BERTopic outperforms other methods like Top2Vec and LDA-BERT in terms of topic diversity and coherence, making it a valuable tool for identifying emerging trends in mental health research. The study also visualizes key machine learning techniques and research trends, underscoring BERTopic's versatility for large-scale data analysis.

Similarly, Ji et al. (2021) (Ji et al., 2021) introduced MentalBERT and MentalRoBERTa, domain-specific masked language models designed to enhance machine learning applications in mental health. By evaluating these models on mental disorder detection benchmarks, the study demonstrated that domain-adapted models significantly improve task performance, emphasizing the importance of pretrained language models tailored for specific domains in mental health research.

While many studies focus on patients' speech or the broader therapist-client interaction, our research shifts the focus to the therapist's language and how their verbal interventions influence the therapeutic process. Using BERTopic, we systematically and scalably analyze therapist remarks, offering new insights into their communication strategies and role in effective therapy.

This study extends the growing body of work on applying machine learning to psychotherapy, contributing to a deeper understanding of the therapist's impact on therapeutic outcomes through their verbal techniques and language patterns.

## 3. Materials and Methods

### 3.1Datasets

The source material was the recordings of psychotherapeutic sessions posted on YouTube in the public domain, which were divided into two datasets: recordings of sessions with classical and modern therapists. The sample of classic psychotherapeutic sessions includes 19 recordings, whereas the sample of modern therapists is of 111 sessions. Classical therapists are represented by recordings of Carl Rogers, Fritz Perls, and Albert Ellis, including their recordings with a client named Gloria. Sessions of modern therapists contain the representatives of various psychotherapeutic directions (Cognitive Behavioral Therapy, Dialectical Behavioral Therapy, Transactional Analysis, Acceptance and Commitment Therapy, Psychoanalysis, Gestalt Therapy, Rational Emotive Behavior Therapy, Motivational Interviewing Therapy, Interpersonal Psychotherapy). After conducting speaker diarization and transcription of the recordings, 8641 utterances were obtained for classical therapists and 4058 for modern ones...

### 3.2Methods

The acquired document corpora were subjected to text preprocessing prior to the investigation, which included segmentation of utterances into separate sentences, performing lexical normalization, cleaning metadata, and converting to a unified case.

Subsequently, topic modeling was applied to the preprocessed corpora using the BERTopic, a machine learning-based topic modeling tool. Pre-trained embeddings from the Sentence-Transformer model 'paraphrase-multilingual-MiniLM-L12-v2' were used to create a vector space. UMAP (Uniform Manifold Approximation and Projection) method was applied to reduce the vector space's dimensionality, and HDBSCAN (Hierarchical Density-Based Spatial Clustering of Applications with Noise) was used to cluster the data. The topic representation of the clusters was made using BERTopic's built-in c-TF-IDF method for assessing the importance of words within the context of document clusters, and its optimization was primarily achieved through the use of large language models like GPT.

After the topic modeling process, the results underwent expert analysis, followed by the removal or merging of topics so as to achieve a more interpretable topic structure of therapist remarks. This required several iterations. At each iteration, interactive visualizations such as hierarchical clustering dendrograms and distance maps across topics were created to aid in the expert evaluation of the modeling results. In addition to visualization, the coherence metric was calculated to assess the topic modeling quality.

At the conclusion of the research, a detailed interpretation of topic clusters was conducted for each corpus of utterances from classical and modern therapists. The most semantically similar topics between two corpora were identified by calculating cosine similarity.

### 3.3 Research Environment and Tools

For this study, we selected Jupyter Notebook version 7.0.6 as our integrated development environment, operating within the Anaconda Python distribution. All data analysis was conducted using the Python programming language.

Libraries and packages specific to each task are as follows:

- Text Preprocessing: The Pandas library (version 2.1.4) was employed for handling tabular data. Text preprocessing was carried out using Python's regular expressions module (re, version 0.12.2) and the advanced natural language processing library spaCy (version 3.7.6);
- Topic Modeling: The BERTopic framework (version 0.16.4) was utilized for identifying topics in therapeutic remarks. Pre-trained embeddings from the sentence-transformers library (version 3.3.1) were employed. Dimensionality reduction was performed using the umap-learn library (version 0.5.7), and clustering was achieved using the hdbscan library (version 0.8.40). Topic coherence was calculated using the CoherenceModel class from the gensim library (version 4.3.3);
- Data Visualization: Matplotlib (version 3.8.0) and Seaborn (version 0.12.2) libraries were used for data visualization. BERTopic's built-in visualization methods were also employed, including hierarchical clustering dendrograms and interactive distance maps between topics based on LDAvis.

## 4. Topic modeling

### 4.1 Data Corpus Preparation

To prepare the corpora for topic modeling, a set of preprocessing operations was carried out using the spaCy library models pretrained on the OntoNotes Release 5.0 corpus (*OntoNotes Release 5.0 - Linguistic Data Consortium*, n.d.) with the additional resources of ClearNLP Constituent-to-Dependency Conversion (*ClearNLP Constituent-to-Dependency Conversion*, n.d.) and WordNet (*WordNet*, n.d.). The following stages were performed during corpora processing:

- segmentation of utterances consisting of splitting the continuous text of utterances into separate sentences in order to isolate syntactic units;
- lexical normalization, which included decoding abbreviations (replacing abbreviated forms of modal verbs and negative particles with their full forms to ensure homogeneity of the vocabulary), unification (removal of uninformative elements, such as interjections, speech fillers and other stop words) and bringing some lexical units to a single orthographic form (for example, "okay");
- cleaning from metadata, which included removing time pause marks (excluding information from the text that indicates the duration of pauses), as well as eliminating identifiers (removing question numbers and other identifiers that do not carry semantic load) in order to preserve only linguistic information;
- bringing to a unified register consisting of converting all letters to lower case so as to unify the text and eliminate the influence of the register on further analysis.

### 4.2 BERTopic Operation Algorithm

Two text data corpora were under topic modeling in parallel using the BERTopic algorithm, which is a multi-stage topic modeling process. The initial stage is the construction of a vector space for each document. A dimensionality reduction technique is used to the resultant vector space in order to increase the effectiveness of further computations and visualization. After that, documents are grouped by topic proximity using a clustering technique. At the final stage, text descriptions of the resulting topics are formed and optimized. A more detailed description of each stage is given below.

#### 4.2.1 Vector Space Construction

We created the vector space from document embeddings obtained using language neural networks with the transformer architecture [4], hence allowing not only to effectively take into account semantic relationships in texts, but also to avoid many stages of text preprocessing, such as removing stop words, stemming or lemmatization (Ma et al., 2021; Thielmann et al., 2023). We applied pre-trained multilingual embeddings of the Sentence-Transformer model 'paraphrase-multilingual-MiniLM-L12-v2' so as to obtain embeddings of the considered text corpora. The choice of this model was due to its higher efficiency compared to others, which was verified empirically, including by comparing with specialized models, for example, BIOBERT, trained on medical texts.

#### 4.2.2 Vector Space Dimensionality Reduction

Since the vector space of embeddings is usually a sparse matrix, BERTopic provides the option to apply a variety of dimensionality reduction algorithms, such as PCA, t-SNE or UMAP (used by default). According to studies (McInnes & Healy, 2017), this step allows expediting the clustering process and increasing its accuracy. Based on the results of (Devlin & Chang, 1810), in our study, we chose the UMAP (Uniform Manifold Approximation and Projection) method with the following hyperparameters:

- number of the local neighborhood points when the manifold approximation, equal to 15;
- dimension of the target space, equal to 5;
- density of points in the low-dimensional space, equal to 0;
- metric used to calculate the distances between points in the original space is Cosine Proximity.

### 4.2.3 Vector Space Clustering

A wide range of clustering algorithms are supported by BERTopic, including contemporary density-based techniques like HDBSCAN and more conventional techniques like k-means and agglomerative clustering. HDBSCAN was used for the current research because it offered the most reliable and accurate text data clustering This algorithm can efficiently handle variable-density data and automatically detect outliers (McInnes & Healy, 2017). We used the Euclidean distance to calculate the proximity between documents in this technique and assumed that a cluster had a minimum of 40 objects.

### 4.2.4 Topic Representation of Clusters

A distinctive feature of the BERTopic algorithm from other topic modeling algorithms (e.g., Top2Vec) is its ability to form semantically rich representations of topics. To do this, a single bag-of-words vector is constructed for each cluster and then a list of the most significant words in the cluster is derived using the c-TF-IDF metric (Kenton & Toutanova, 2019). This metric is an extension of traditional TF-IDF adapted for cluster analysis and allows assessing the importance of terms (words) in the context of a specific cluster.

Determining the importance of terms makes it possible to create a list of the most relevant keywords describing a certain cluster, which is the topic representation of the cluster in question. This opens up options for optimizing the cluster structure by combining topics with similar semantic content, hence reducing the number of clusters and minimize the number of outliers.

In addition to a relative accurate representation of clusters, BERTopic enables the optimization of topic representation by using built-in additional techniques. These include methods based on keywords (KeyBERTInspired), maximum marginal relevance (MMR), part-of-speech analysis (using the spaCy library), and the application of large language models (e.g., GPT or T5). The latter are particularly promising due to their ability to automate the process of generating cluster descriptions and significantly lessen the workload of experts.

We combined all of the aforementioned techniques in our study so as to further personalize the topic representations. In order to generate a description of each cluster, we used the following prompt to the GPT model:

```
"""
I have a topic that contains the following documents:
[DOCUMENTS]
The topic is described by the following keywords:
[KEYWORDS]
Based on the information above, extract a short but highly
descriptive topic label of at most 5 words. Make sure it is in
the following format:
topic: <topic label>
Topic must be in the language in which the documents are
written
"""
```

## 4.3 Assessment and Optimization of Modeling Outcomes

Applying the BERTopic topic modeling algorithm to two text corpora containing therapists' remarks from different eras yielded 95 and 110 topic clusters, respectively. The semantic similarity between words in topics, estimated using the topic coherence metric (Mimno et al., 2011), was 0.429 and 0.377, respectively. Interactive visualization of topic spaces using LDAvis (Sievert & Shirley, 2014) (Fig. 1) validated the results, demonstrating the degree of semantic similarity between specific topics in each corpus.

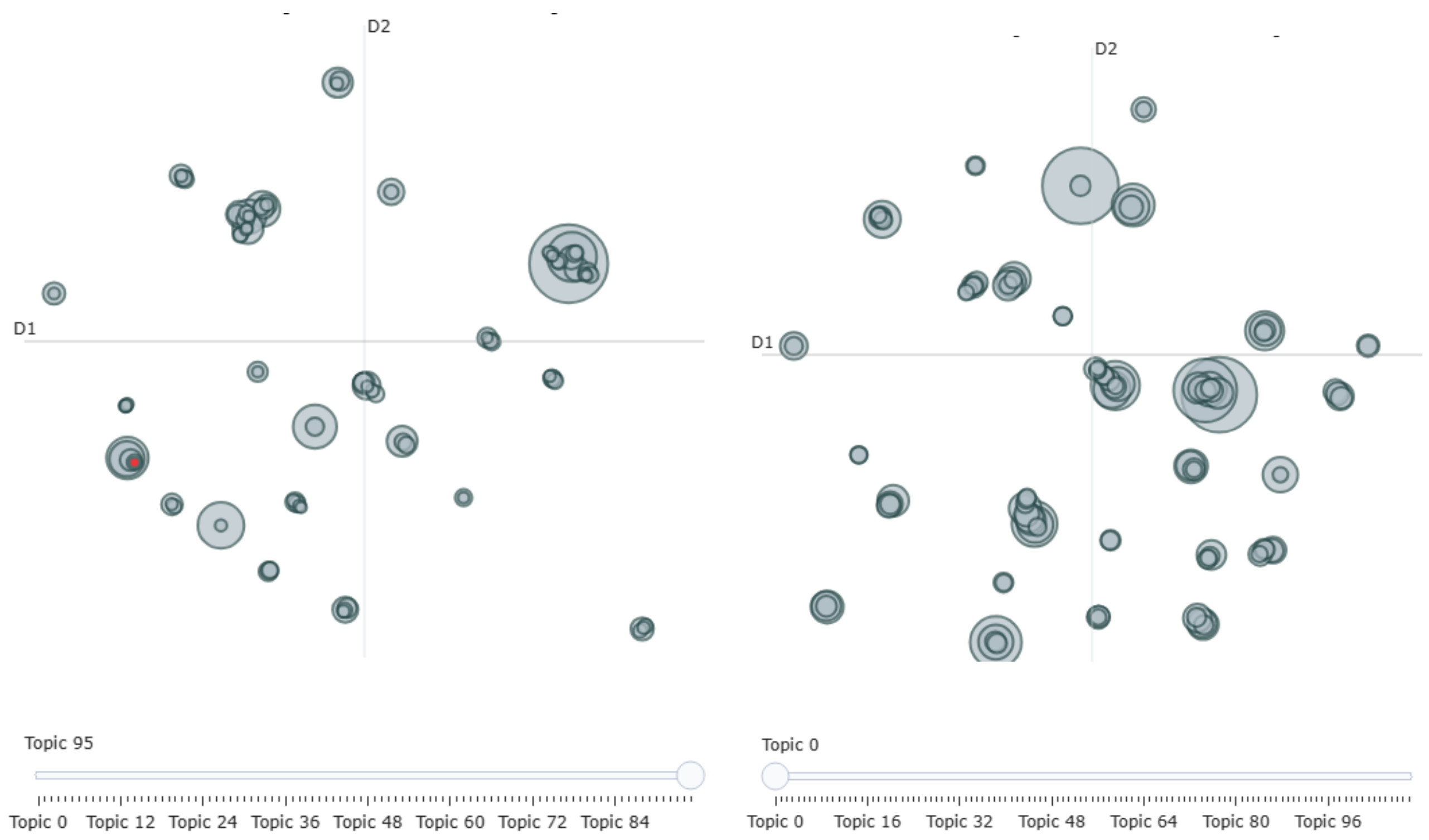

(a) (b)

**Fig. 1** Maps of distance between topics for:
a) utterances of classical therapists, b) utterances of modern therapists

Through expert interpretation of the topic modeling results, we were able to identify a large number of similar clusters in terms of the subjects' meanings. These clusters were either classified as close in cosine distance by the BERTopic model or distant, but still having similar semantic content by the expert view. In both cases, these topics displayed a sufficient number of representative documents to be considered separate clusters (as indicated above, the minimum number of documents in a cluster was taken to be 40). So as to combine the clusters we applied the merge_topics method (built into the model) creating a new cluster by averaging the vectors of the merged topics. In addition, the clusters consisting of filler words and discourse markers with little semantic weight (such "yes," "Uhm-hm," "Is that...?" etc.) were also merged into a distinct category called "Others".

As a result of manual refinement of thematic structures based on expert knowledge, 43 and 46 clusters were identified in the utterances of therapists for classical and modern directions, respectively. At the same time, the coherence of topics increased, as expected, to 0.47 and 0.431, respectively. Below is a thorough analysis of the clusters' content.

## 5. Interpretation of Topic Clusters

To present the results, we will examine the identified topics in the therapists' remarks sequentially, referencing their representation in the results of hierarchical cluster analysis (Fig. 2-3), which determines the proximity of topics to each other. We will begin by discussing the topics associated with classical therapists, followed by an exploration of the topics associated with modern therapists. Finally, we will compare the proximity of topics between classical and modern therapists (Fig. 4).

### 5.1Topic Structure for Classical Therapists

For classical therapists, 44 topics were identified, which are described in detail below. These topics represent the various speech patterns, techniques, and themes that emerged in their dialogues.

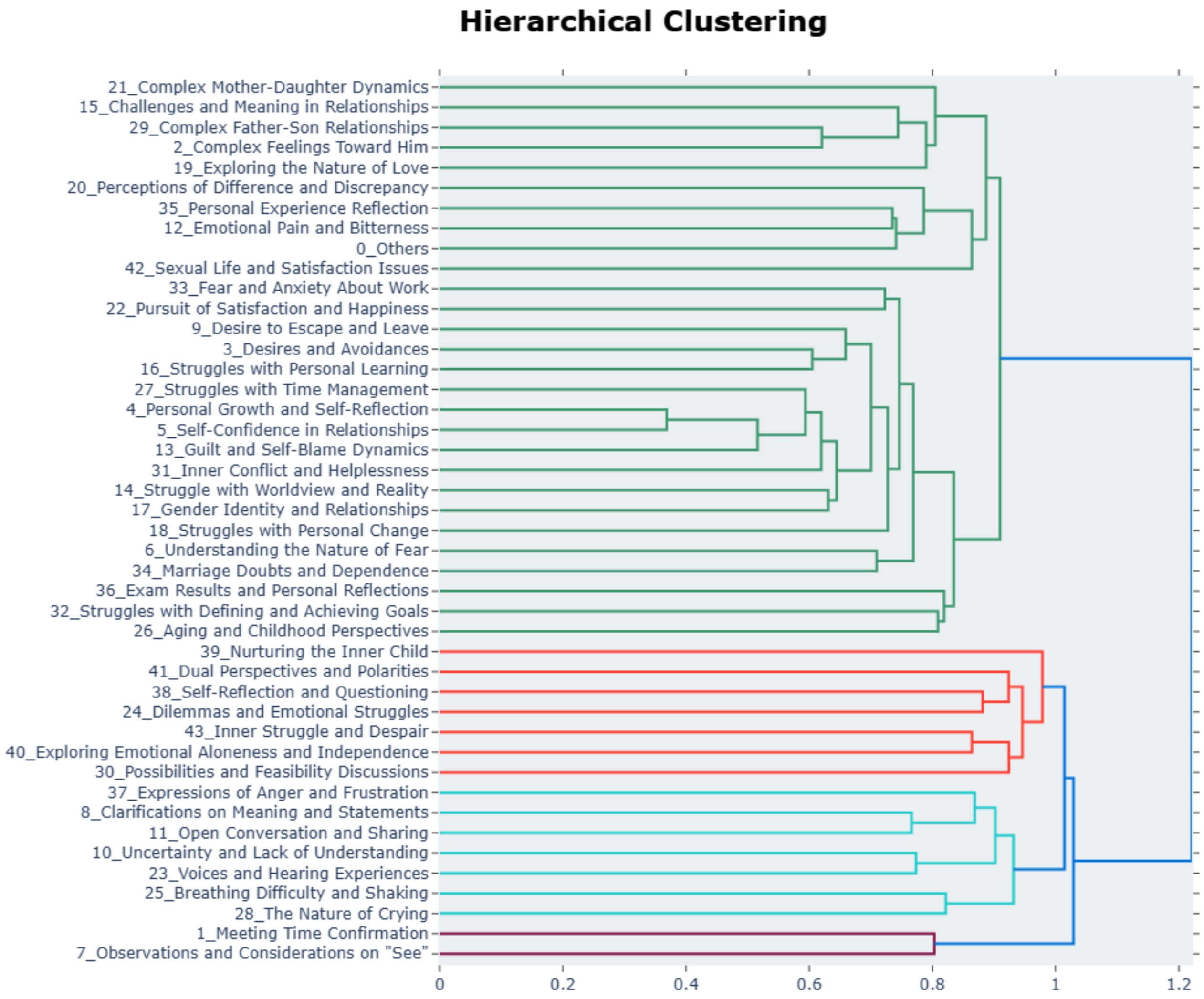

**Fig. 2** The results of hierarchical cluster analysis of the classical therapist's statement topics

Order of consideration – **(1)**, Topic number – **21**, Topic title – **Mother-Daughter Relationship Dynamics.** Keywords (‘K’): ['mother', 'your mother', 'your', 'mother and', 'my mother', 'my', 'to', 'you', 'her', 'that']. Number of texts (‘N’): 61. The therapist (‘T’) analyzes the client's (‘C’) relationship with her mother («*It makes you feel as though…you are a mouse in your relationship to your mother*») and notes C’s personality traits that are similar to those of her mother («*So, you’re saying your mother was the model for this organizing self*»).

**(2) 15 Complexities of Relationships.** K: ['relationship', 'the relationship', 'relationships', 'relationship that', 'relationship you', 'that relationship', 'of', 'the', 'that', 'like']. N: 78. T analyzes various aspects of C’s relationships with others («*You know that it’s a difficult relationship, and you’d like to preserve it, but it’s going to take a great deal of energy on your part to hold that relationship together*»), including possible strategies for improvement («*It doesn't spoil the relationship to say ‘No’*»).

**(3) 29 Father-Child Relationships and Authority**. K: ['father', 'your', 'your father', 'dad', 'with', 'with your', 'he', 'stepfather', 'relationship', 'brother']. N: 47. T analyzes various facets of C’s relationship with their father or stepfather («*That things went somewhat better between you and your father*», «*Just like your father*»).

**(4) 2 Complex Emotions Toward Him.** K: ['he', 'him', 'his', 'that he', 'you', 'that', 'to', 'and', 'feel', 'not']. N: 302. T examines C’s relationship with an abstract male figure («*You feel he has a dark critic inside of him*», «*There are certain things you respect him for, but that doesn’t, uh, alter the fact that you definitely hate him and don't love him*»).

**(5) 19 Exploring Love and Self-Acceptance**. K: ['love', 'person', 'somebody', 'to', 'want', 'you meet', 'yourself', 'now let', 'meet', 'affection']. N: 63. T analyzes C’s romantic relationships and self-acceptance. For instance, T normalizes C’s feelings regarding unrequited love («*So, it hurt not to get that love back*») or explores the facets of love («*That there is a need for a love that isn't possessive and that just exists*»).

**(6) 20 Perceived Differences and Contradictions.** K: ['different', 'something', 'is something', 'contradictory', 'it', 'is', 'difference', 'that', 'different from', 'discrepancy']. N: 63 T addresses inconsistency between people's words and behavior («*It is just an absurdly contradictory situation, in which we people say one thing and do the complete opposite*»). T also explores differences among individuals («*They somehow seem to belong together and yet they came out quite differently*»), including differences in personal values («*It sounds as though it isn't any one big difference in values but a number of little things that pile up and pile up until you think, 'Oh, the difference is too great'*»).

**(7) 35 Personal Experience Reflection.** K: ['experience', 'own experience', 'experienced', 'about', 'it', 'situation', 'about it', 'that', 'experience that', 'you have']. N: 30. T encourages C by highlighting their positive life experiences («*It really was pretty swell the way you handled that fraternity situation on the photographic job*») and also refers to their own experience («*I'll simply speak from my own experience, and I'm not sure that it applies to you*»).

**(8) 12 Exploring Hurt and Bitterness.** K: ['hurt', 'pain', 'bitterness', 'the hurt', 'of', 'and', 'it', 'hurt and', 'the bitterness', 'the']. N: 85. T explores the area of C's emotional pain and bitterness. Specifically, T identifies the presence of painful feelings («*It looks like you're feeling some of that hurt right now*»), analyzes their causes («*That's the way you have been hurt*»), emphasizes the importance of reflection on pain («*It's been so good to get out both the hurt and the bitterness where you can look at them*»), and explains the different facets of pain to C («*There's a lot of pain there – pain about the past and pain about the present*»).

**(9) 0 Others.** K: ['mhm', 'mhm mhm', 'her', 'she', 'that', 'the', 'okay', 'mhm that', 'not', 'it']. N: 3827. This category includes various topics, as well as paralinguistic expressions, which were removed due to their insufficient depth for analysis.

**(10) 42 Sexual Satisfaction and Relationships.** K: ['sexual', 'sex', 'life', 'normal', 'satisfaction', 'that the', 'that', 'of', 'maybe', 'and']. N: 21. T analyzes various aspects of C's sexuality, including C's satisfaction with their sexual life («*If you couldn't find sexual satisfaction, that would be a failure of a very deep sort*»), psychoeducation on this topic («*And it may say in the books that with no sex outlet, you would have to be neurotic*»), and sexual frustration («*That you can say that the base of all is the sexual frustration and so forth*»).

**(11) 33 Fear and Anxiety About Work.** K: ['job', 'work', 'the', 'to work', 'you', 'get job', 'job or', 'do', 'the job', 'of work']. N: 36. T discusses C's negative emotions related to work («*You feel as though staying in the job situation may really bring a blow-up on your part*», «*That is, sort of asking yourself, why the hell should I be fearful of or overwhelmed by the notion of a job?*»).

**(12) 22 Pursuit of Personal Satisfaction.** K: ['satisfaction', 'satisfactions', 'happiness', 'the satisfactions', 'the', 'more', 'of', 'you', 'good', 'of satisfaction']. N: 56. T addresses C's sense of inner satisfaction («*It seems more that the sense of satisfaction is a sort of total process of change that you feel and you move with it*») and seeks to offer C an alternative perspective on the possibility of achieving this state («*You realize there could be some satisfaction in doing something different from what you might be told*»).

**(13) 9 Desire to Escape and Leave.** K: ['away', 'away from', 'leave', 'it', 'you', 'from', 'to', 'out', 'get', 'get away']. N: 118. T explores both the C's desire to escape the situation they find themselves in («*Seems as though the thing that is stirring within you is more 'I want to get away, be out on my own, be free'*») and the difficulties that accompany this («*And you know what a struggle it was to move away from that*», «*Just felt that there was a need to get away and you couldn't explain it to others; you just couldn't*»).

**(14) 3 Desires and Reluctance.** K: ['do', 'not', 'do not', 'want', 'not want', 'you do', 'want to', 'you', 'to', 'anything']. N: 274. T identifies desirable and undesirable scenarios for C («*You don't want to make it any darker than it is*», «*You don't want to starve and you don't want to cross the Alps*»).

**(15) 16 Personal Journey of Learning.** K: ['learn', 'to learn', 'learning', 'you', 'school', 'to', 'and', 'your', 'the', 'what']. N: 71. T explores C's desires in the areas of learning and education («*You feel a real desire to try to learn some of this for what you would like to learn from it, not in terms of meeting some future examination demand or any other future demand*», «*You feel that you really like to achieve and learn a lot, but you just, you're just not doing it at present*»).

**(16) 27 Struggles with Time Management.** K: ['time', 'your time', 'you', 'to', 'that', 'not', 'just', 'the', 'feel', 'you feel']. N: 49. T addresses C's experience of life's meaningfulness («*In other words, it isn't just a question of filling in your time...*»), their sense of time scarcity («*You feel that right now you really waste a great deal of time that you actually can't afford to waste*»), and aspects of time management («*You've always felt that there was hardly enough time and you've had to utilize every scrap of it to get your things done*»).

**(17) 4 Personal Growth and Challenges.** K: ['you', 'to', 'that', 'the', 'of', 'in', 'that you', 'and', 'it', 'feel']. N: 215.

T addresses C's lack of clarity regarding their life direction («*You don't know at all where you want to, what direction you want to move or what you want to do*»). T also interprets C's indecision in choosing a life path («*You feel that on the one hand you're not living up to all the things you really should be doing, but on the other hand, you feel that it's more realistic to grow gradually into that*») and highlights their progress in personal growth («*I feel that you've made a good deal of progress inside yourself*»).

**(18) 5 Self-Acceptance and Confidence Issues.** K: ['you', 'that', 'in', 'feel', 'you feel', 'to', 'confidence', 'them', 'your', 'the']. N: 212. T explores C's confidence and lack of confidence in themselves. For instance, T notes C's confidence in their feelings («*You feel kind of a confidence in your own feelings*»), discusses potential reasons for C's sense of inadequacy («*You feel that something should have turned up to give you that confidence in yourself*»), and analyzes aspects of C's self-perception in the context of how they are perceived by others («*You feel that you are living by the standards others have and what they think of you and so on, even though more deeply you know that you can't possibly have happiness that way*»).

**(19) 13 Guilt and Self-Blame Dynamics.** K: ['blame', 'for', 'you', 'feel', 'to blame', 'that', 'yourself', 'guilt', 'you feel', 'guilty']. N: 80. T addresses C's sense of guilt («*You feel guilty about what you haven't done, and it gives you more reason for avoiding the people*»), tries to explain the origins of this feeling («*You don't like to blame your school or your family, but still you feel that to some extent your family was responsible*»), and explores C's self-blame («*That sounds like not only your family looks down on it or something but that you scold yourself for it too*»).

**(20) 31 Inner Struggle and Helplessness.** K: ['struggle', 'you', 'it', 'you feel', 'pretty', 'feel', 'that', 'and', 'situation', 'to']. N: 46. T explores C's inner struggle («*You're inclined to feel that war situation or no war situation, the struggle is pretty much within you, after all*»), their sense of helplessness («*You feel that it's really the back of the coming in is the fact that you feel helpless to do anything about it*»), potential strategies for overcoming this feeling («*In other words, when you begin to feel hopeless then it seems so necessary to distract yourself from yourself*»), and the possibility of external support («*It's pretty deeply annoying to get into the conflicts and then not be sure which way to go and wish like hell somebody would give a little push*»).

**(21) 14 Intellectual Disconnection and Existence.** K: ['world', 'intellectual', 'the world', 'the', 'it', 'you', 'in', 'not', 'universe', 'of']. N: 79. T explores C's relationship with the world around them («*You feel at real odds somehow with the world in general*», «*You don't like that feeling that the whole world's wrong*»). T also legitimizes C's value orientations («*I don't know whether it's as a philosopher, but I certainly would agree with you that, in situations of this kind, I don't think there is any proof that could be advanced that would prove one set of values rather than the other*») and challenges excessive rationalization («*So maybe the intellectual understanding isn't as important as you thought it was*»).

**(22) 17 Gender Roles and Relationships.** K: ['women', 'woman', 'men', 'feminine', 'you', 'man', 'role', 'in', 'of', 'that']. N: 68. T examines C's gender roles. Specifically, T interprets C's issues («*So, it's really a big problem for you, you are feminine and you like to be feminine and you're seen as feminine and reacted to in that way, and then you think, 'Oh my God!'*», «*When you feel insecure about the better sexual relationship, it might be because it hits you kind of hard that that stresses the fact that you are a woman with a woman's needs*») and notes C's masculinity («*You see both elements of yourself pretty sharply, where on the one hand you may be more of a woman than you think you are, and on the other hand you have pretty masculine interests along some lines*»).

**(23) 18 Struggles with Personal Change.** K: ['change', 'to change', 'changes', 'past', 'you', 'to', 'change in', 'the', 'in the', 'in']. N: 67. T addresses C's personal changes: touching on aspects such as C's procrastination («*You think to make a change you ought to change right now; on the other hand, you think that maybe it would have been simpler to delay and change next quarter*») and interpreting the difficulties with change that C experiences («*But you're just very much aware of a loosening type of change taking place, and intellectually when you can stand off and look at it to see, if...you knew exactly what made the wheels go round, hm?*», «*You tend to dwell a little more on what you didn't do in the past, rather than on what might be done right now or in the future*»).

**(24) 6 Exploring the Nature of Fear.** K: ['fear', 'afraid', 'the fear', 'of', 'of fear', 'the', 'fear of', 'afraid of', 'you', 'risk']. N: 200. T addresses the nature of C's fear, including various forms of fear («*They nevertheless are fears deep inside, and the biggest fear of all is the fear of being trapped, in so many different ways*», «*So that's where you fear is, a fear of a relationship with a man*», «*It really wasn't the fear of death; that you can accept*», «*Fear of choking, fear of convulsions?*», «*It seems to me that you're saying, ahh, the fear, the fears grow stronger,* *as time goes by, both of marriage and of children and of commitment, as well as the fear of aging... that it seems a package of fears*»). T also attempts to normalize C's fear («*The fear comes and goes*», «*To know that you are not alone in that fear, the others have the same kind of fear*») and interprets their anxious state («*The more clear our feeling of fear that exists in us here is blocked off close to you, that then somehow it develops into this more nameless, more indescribable kind of fear that is anxiety*»).

**(25) 34 Marriage Doubts and Dependence.** K: ['marriage', 'husband', 'your husband', 'married', 'your', 'your marriage', 'on', 'trapped', 'you', 'the']. N: 34. T examines C's relationship with their life partner, including aspects such as C's feelings in marriage («*You're not happy with your husband and he's not happy with you?*») and fear of commitment («*There is a fear of commitment, and a fear of having children. And a feeling that in marriage you don't want to give up your identity*»). T also presents for consideration C's hidden reasons for entering into marriage («*Perhaps one of the things you looked forward to in marriage was that there would be a situation and a person on which you could basically depend*»).

**(26) 36 Exam Results and Personal Reflection.** K: ['test', 'exams', 'exam', 'the exam', 'the results', 'results', 'the', 'tell you', 'do about', 'what to']. N: 30. T discusses both the exam situation with C («*At least you're not sorry for having taken the exams and having made that decision*») and educates C about existing psychological tests («*And the Rorschach test you probably know – that's the one with the ink blots*»), providing their results («*Well, I can show you the results of the tests*»).

**(27) 32 Understanding Personal Goals and Satisfaction.** K: ['goal', 'goals', 'the goal', 'reach', 'you', 'the', 'and', 'that', 'satisfied', 'achieve']. N: 45. Next, T examines C's personal goals. T analyzes the nature of these goals («*In other words, the goal as nearly as you can formulate it is some kind of fusion between the things intellectually you know you want and something pretty deep in you that doesn't lend itself easily to words, or labels or...*»), notes the connection between goal achievement and self-knowledge («*Whereas the kind of goal you want to reach probably can be more easily achieved when you understand some aspects of yourself*»), assesses C's progress in reaching personal goals («*It's just the gradual realization that you are not as far toward the goal as you hoped*»), and identifies the necessary conditions for C to achieve these goals («*You expected to reach the goal without the work or struggle that went in between*»).

**(28) 26 Aging and Childhood Reflection.** K: ['childhood', 'childish', 'age', 'old', 'eighteen', 'young', 'you', 'and', 'childhood and', 'your childhood']. N: 50. T explores C's world through the lens of childhood and aging («*You may behave in a certain way, and then only later realize that that was a step from childhood into adolescence*»). T also addresses C's fear of growing old («*Can you tell me a little bit more about your fear that you have of aging?*»).

**(29) 39 Embracing the Inner Child.** K: ['girl', 'little girl', 'little', 'naughty little', 'naughty', 'daughter', 'daughter you', 'as my', 'the little', 'that little']. N: 24. T metaphorically and with a touch of irony addresses C, emphasizing her attention on her inner self («*You know that little girl is inside of you*», «*The naughty little girl can get away with things*») and also addressing the nature of growing up («*The little girl, the little girl will grow up if you care enough for her*»).

**(30) 41 Dual Perspectives and Polarities.** K: ['two', 'sides', 'both', 'one', 'so there', 'two sides', 'split', 'both sides', 'angle of', 'are two']. N: 23. T addresses the duality of C's nature («*Between the two selves*») and desires («*You want both of those directions*»), exploring their polarization («*But I get a sense of two poles there*»).

**(31) 38 Self-Reflection and Questioning.** K: ['question', 'the question', 'asking', 'asking yourself', 'raising', 'question you', 'yourself', 'question that', 'yourself so', 'you are']. N: 25. T notes C's self-reflection («*It sounds like you're asking yourself that, as well as me*», «*So, you're raising the whole question, 'Is it really productive to discuss so much about these value systems?'*») and the frequent repetition of the same questions by C to themselves («*That's the question you keep asking yourself*»).

**(32) 24 Handling Difficult Dilemmas and Emotions.** K: ['too', 'too much', 'too strongly', 'hard', 'tough', 'putting it', 'dilemma', 'difficult', 'strongly', 'it']. N: 52. T addresses the dilemmas and difficulties in C's life, possibly with the intention of normalizing their emotions regarding the situation they find themselves in («*It is a real dilemma*», «*For you, you feel it's almost too much, is that it?*»).

**(33) 43 Struggles with Self-Perception.** K: ['worse', 'dirty', 'you', 'it', 'felt', 'you as', 'awful', 'just', 'within yourself', 'felt that']. N: 20. T addresses the negative aspects of C's self-perception («*Intellectually it might seem as though maybe you should find something very awful within yourself*», «*That everything inside of you is worse than you thought*», «*You really felt you had gotten a very dirty deal from nature*»).

**(34) 40 Emotional Aloneness and Independence.** K: ['alone', 'let me', 'alone you', 'aloneness', 'to alone',

'lonely', 'because do', 'with you', 'independent', 'as though']. N: 24. T examines C's sense of loneliness («*You felt very much alone emotionally*», «*It's as though you're in some way sort of responsible for your loneliness*»).

**(35) 30 Possibilities and Potential Outcomes.** K: ['possible', 'might', 'that would', 'would', 'would that', 'maybe', 'that', 'perhaps', 'maybe that', 'possibility']. N: 47. T draws C's attention to various possible outcomes («*It's one possible option at any rate*», «*Yes, probably that would be another possible outcome*»).

**(36) 37 Understanding and Expressing Anger.** K: ['anger', 'angry', 'rage', 'anger and', 'and', 'there', 'you could', 'explaining', 'angry at', 'the anger']. N: 27. T explores the nature of C's anger. Specifically, T notes anger in C's speech («*I can hear the anger in there*»), legitimizes C's desire to feel anger («*If you feel like being angry, you can be angry*»), and discusses anger management strategies («*So, when you meet anger, whether in men or women, you tend to placate it if possible*»).

**(37) 8 Clarifying Meaning in Conversations.** K: ['what you', 'saying', 'that what', 'what', 'are saying', 'say', 'mean', 'meaning', 'you mean', 'to say']. N: 157. The topic contains T's statements aimed at clarifying what C has said («*Is that what you mean?*», «*Is that what you're saying?*»).

**(38) 11 Open Conversations About Feelings.** K: ['tell me', 'tell', 'me', 'talk', 'appreciate', 'to talk', 'what', 'about', 'to tell', 'to']. N: 99. The topic contains T's statements intended to begin the therapy session with C («*Now, what I would like would be for you to tell me anything you're willing to tell me about yourself and your situation*»), as well as other requests for information from T to C («*Tell me, what's your first name?*», «*Can you tell me any more of your thinking about it?*»).

**(39) 10 Uncertainty and Clarity Issues.** K: ['not know', 'not quite', 'sure', 'not', 'quite', 'am not', 'do not', 'know', 'quite sure', 'am']. N: 105. T acknowledges that they do not fully understand C or their situation («*I'm not quite sure*»), and also notes the uncertainty in C's words («*You're not quite sure why*»).

**(40) 23 Voice Perception and Despair.** K: ['voice', 'hear', 'the voice', 'did not', 'heard', 'did', 'not hear', 'hear you', 'not', 'quite']. N: 52. T metaphorically addresses C's inner voice («*I guess another puzzling thing about it is that you feel that you hadn't heard the voice for quite a while, why, why would it return?*», «*And was it the voice that said, 'If you are feeling desperate...'*»), also noting the despair in C's tone («*Oh, the voice sounded kind of desperate*»). This topic also includes T's statements about the volume of C's voice («*I didn't quite hear what you said*»).

**(41) 25 Breathing Struggles and Shaking.** K: ['shaking', 'can not', 'throat', 'choking', 'head', 'your', 'your throat', 'choke', 'can', 'place of']. N: 51. T interprets the cause of C's trembling («*Your shaking is really fear*») and also notes other physiological manifestations in C's behavior («*You can't breathe*», «*Your head's spinning?*»). This topic also includes C's smoking habits («*It doesn't keep you from going back to smoking*», «*I don't smoke, but if you've got cigarettes, feel free to light up one*»).

**(42) 28 Crying and Emotional Release.** K: ['cry', 'tears', 'the tears', 'if', 'crying', 'brings', 'would', 'weeping', 'cry it', 'brings the']. N: 48. T discusses sensitive topics with C that may evoke tears («*It almost brings tears to your eyes, doesn't it?*»), also noting C's tears («*There are tears now*») and normalizing their feelings about them, thereby expressing empathy («*Yes, you could cry*»).

**(43) 1 Time Management and Scheduling.** K: ['we', 'time', 'our', 'our time', 'week', 'next', 'see', 'monday', 'friday', 'stop']. N: 355. The topic includes both T's statements regarding the management of the therapy session's duration («*I see our time is up*») and remarks directed at planning a new meeting («*Shall we make it next week at the same time?*»).

**(44) 7 Understanding the Concept of Seeing.** K: ['see', 'see see', 'mhm see', 'see mhm', 'let', 'let is', 'get that', 'get', 'is see', 'see if']. N: 175. T demonstrates their readiness to address the problematic situation («*Let's see*», «*But let's just look at that*») and shows understanding in response to C's remarks («*I see*»), thereby supporting the dynamics of the therapeutic process.

## 5.2Topic Structure for Modern Therapists

For modern therapists, 47 topics were identified, which are described in detail below.

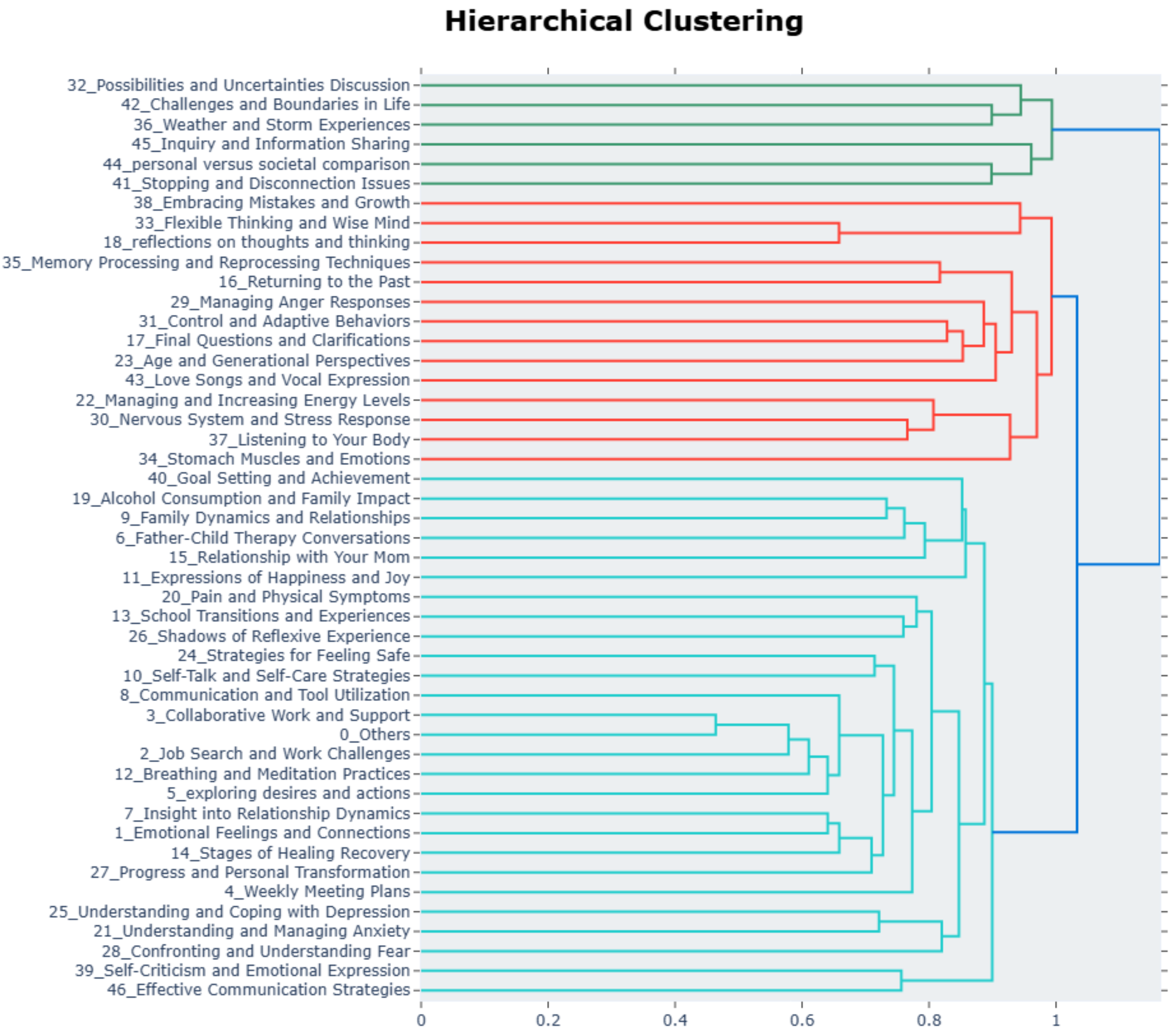

**Fig. 3** The results of hierarchical cluster analysis of the modern therapist's statement topics

**(1) 32 Possibilities and Uncertainties.** K: ['possibility', 'maybe', 'probably', 'might', 'well', 'is possibility', 'well maybe', 'well it', 'it could', 'possible']. N: 32. T considers various scenarios for the development of events in C's situation («*Right, that's one possibility*», «*Another possibility*»).

**(2) 42 Navigating Difficult Situations.** K: ['difficult', 'hard', 'it is', 'easy', 'it', 'an easy', 'not an', 'very', 'is', 'boundaries']. N: 23. T identifies the boundaries of the complexity of C's situation («*Yes, so it's a little more difficult than that*»), normalizes C's feelings regarding the situation's difficulty («*It's a difficult thing to pull off*»), and motivates C («*Again, even if it is hard*»).

**(3) 36 Weather and Storm Descriptions.** K: ['weather', 'wet', 'the weather', 'storm', 'water', 'rain', 'rihanna see', 'rihanna', 'today', 'mean']. N: 30. T uses weather and storms as metaphors to describe C's mood («*How's the weather relating to you then?*», «*All I would want you to do is to remember the storm has a beginning, a middle, and an end*»).

**(4) 45 Inquiry and Information Seeking.** K: ['tell me', 'tell', 'me', 'about that', 'about', 'so tell', 'to know', 'let', 'know about']. N: 20. T maintains a dialogue with C and requests information from them («*Yeah, so let's talk about that*», «*So, tell me a little bit about the…*»).

**(5) 44 Personal vs. Societal Comparison.** K: ['comparison', 'comparing', 'compare', 'personal', 'more of', 'the comparison', 'motivation', 'to compare', 'why', 'though you']. N: 21. T discusses a topic such as C's motivation in comparison to other people («*And so, if you would like to stop comparing and we know the function of the comparison part of it and she is helping you it is effectively and that the motivation of you doing comparison*», «*Comparison because you're competitive *you want to make sure that you are moving along with the rest and not getting lucky*»).

**(6) 41 Processes of Stopping and Disconnecting.** K: ['stop', 'shut', 'disconnect', 'stopped', 'stop the', 'down', 'shut down', 'because think', 'it down', 'because']. N: 27. T explains the concept of stopping and disconnecting to C («*This whole procedure is called stop*»), manages the therapy session («*Stop for a little bit*», «*Yeah, we probably should stop*»), and suggests that C learn to pause their emotions in certain situations («*Stop the feeling that automatically starts*»).

**(7) 38 Understanding Mistakes and Consequences.** K: ['mistake', 'mistakes', 'make mistake', 'make', 'consistent', 'in thinking', 'to make', 'thinking', 'everybody', 'they are']. N: 29. T discusses the topic of mistakes, including cognitive errors («*They're consistent errors in thinking*»). T also normalizes C's feelings («*Everybody makes mistakes*», «*It's reasonable that a professional makes a mistake like that*»).

**(8) 33 Flexible Thinking and Mindfulness.** K: ['thinking', 'mind', 'logical', 'of', 'thoughts', 'logical mind', 'wise mind', 'those thoughts', 'might', 'that']. N: 31. T analyzes the thinking process with C («*Before you didn't think to think about the thought that makes sense and there are a few steps nearby*»), examining the influence of thoughts on feelings and behavior («*But you could have those thoughts working for you*»), as well as the process of evaluating cognitive acts («*I'll learn how to evaluate my thinking, which might be a hundred percent true, or 0% true, or someplace in the middle*»).

**(9) 18 Thoughts and Mind Exploration.** K: ['thought', 'think about', 'about', 'thinking', 'mind', 'think', 'thoughts', 'your mind', 'thinking about', 'what']. N: 74. T asks C what they are thinking, both in general and in relation to the situation («*What was going through your mind?*», «*What comes to you when you think about it?*»), and also guides C to analyze the situation («*Kind of thinking about that thought and what happens afterward*»).

**(10) 35 Memory Processing and Reprocessing.** K: ['memory', 'memories', 'target', 'bring up', 'the', 'up', 'the target', 'we', 'past', 'bring']. N: 30. T refers to C's memories («*And I want you to think back to what is kind of like the earliest memory that you have of having that feeling or thought of I need to be in control*») and also describes the process of memory work («*And so, what that does is it really opens up the memory capsule or the memory storage and then that processing phase starts and what the reprocessing stage is essentially just to demystify; all it means is you're going to bring up all that material*»).

**(11) 16 Returning to the Past.** K: ['back', 'go back', 'back to', 'go', 'history', 'past', 'back into', 'to the', 'back and', 'so']. N: 77. T brings C back to a previously discussed issue («*So let us go back to what we had before*») or asks C what they would do if they could return to a specific point in the past («*If you could go back and…*»).

**(12) 29 Managing and Understanding Anger.** K: ['angry', 'anger', 'your anger', 'anger and', 'consequence', 'you are', 'angry and', 'are', 'you', 'of']. N: 43. T analyzes C's anger response («*What types of things do you think are triggering your anger response?*»), the impact of anger on thinking and behavior («*And it's hard when you're angry because when your anger is your focus on…*»), and the possibilities for managing anger («*In the consequence of that thought is I'm I'll be angry but I'm not going to act angry*»).

**(13) 31 Control and Behavioral Strategies.** K: ['control', 'can', 'behavioral', 'behaviors', 'behavior', 'behavior so', 'not control', 'can not', 'we', 'think']. N: 32. T discusses with C strategies and techniques for self-regulation of behavior («*So that would be one behavioral strategy*»), establishes boundaries of what can be controlled for C («*We can only control what we're dealing with right here*»), and engages in a conversation about the dichotomy of control («*Do you think all three of those instances extend from the same thing of that something that I can't control like these things are outside of my control?*»).

**(14) 17 Final Questions and Clarifications.** K: ['questions', 'any', 'any questions', 'quiz', 'question', 'of questions', 'final questions', 'any final', 'any any', 'questions you']. N: 75. T maintains the dialogue by asking C if they have any questions: before the start of the therapy session («*All right, so I guess before we get started, do you have any questions for me?*»), during it («*You can talk about any questions*»), and before its conclusion («*Did any final questions that you got for now?*»).

**(15) 23 Age and Generational Perspectives.** K: ['old', 'how old', 'generations', 'old you', 'age', 'how', 'teenager', 'younger', 'you', 'young']. N: 57. T asks about C's current age («*How old are you?*») and also inquires about events in C's life in relation to the age they were at the time those events occurred («*How old were you when that started?», «So, you were how old when he passed away?*»).

**(16) 43 Love Songs and Voice Exploration.** K: ['song', 'voice', 'can', 'songs', 'sang', 'that song', 'hear you', 'hear', 'can you', 'this to']. N: 21. T says that they hear C («*I can hear you*»), asks if C hears them («*Tatiana, can you hear me?*»), and also reflects on the topic of songs («*Love songs that kind of inspiration YouTube thing at school and chemistry and just working stuff out*»).

**(17) 22 Managing and Increasing Energy.** K: ['energy', 'the energy', 'the', 'of energy', 'of', 'and', 'your', 'more energy', 'more', 'you']. N: 60. T discusses C's energy («*You're increasing your capacity that will bring more of your energy into your body*») and notes that it needs to be managed effectively («*Start investing a lot of energy*», «*That's what you need to do as the energy comes back fast; a real challenge is what to do with that energy at that point*»).

**(18) 30 Nervous System and Stress Response.** K: ['nervous system', 'nervous', 'system', 'your nervous', 'stress', 'your', 'of', 'kind', 'stress response', 'kind of']. N: 41. T discusses C's nervous system («*Do you know that that makes your nervous system?*»), including its connection to emotions and stress («*Do you feel like you're in a stressed-out overheating?*»), and how to cope with them («*My point is you want your nervous system to not be so agitated that you need something like to switch it off in the first place*»).

**(19) 37 Listening to Your Body.** K: ['body', 'your body', 'your', 'body and', 'and', 'the body', 'legs', 'in', 'listen to', 'arms and']. N: 29. T notes the connection between the body and the mind («*It's a way for your body trying to get your attention*») and encourages C to listen to their body («*I just really encourage you to listen to your body and work on your relationship with your body and soul*»).

**(20) 34 Stomach Feelings and Tension.** K: ['stomach', 'your stomach', 'your', 'side of', 'in your', 'chest', 'side', 'tool', 'throat', 'your chest']. N: 31. T works with C's physiology, such as the abdominal muscles, to relieve tension («*Let's see what it says your stomach muscles in your chest when you're angry*», «*Clench your stomach*»).

**(21) 40 Goal Setting and Achievement.** K: ['goal', 'goals', 'your goal', 'oriented', 'goal for', 'your', 'your goals', 'you', 'reach', 'for']. N: 27. T discusses C's personal goals («*You have a goal that you're trying to reach because your goal is to be a doctor, right?*»), including those related to a specific therapy session («*So, what would be your goal for today's session?*») and therapy as a whole («*What would be your goal here?*»). T also talks about the goal achievement process («*I think that's important for us to note that planning is absolutely a part of what it would take for you to reach your goals*»).

**(22) 19 Alcohol Consumption and Consequences.** K: ['drinking', 'alcohol', 'drink', 'you are', 'you', 'are', 'your', 'the', 'so', 'about']. N: 73. T discusses C's alcohol consumption («*Okay, so to your wife, you're drinking is excessive, but to you, it is pretty normal*»), the reasons behind it («*The drinking has become a way for you to unwind and release some of the stress that you are feeling at work*»), its impact on C («*But when you're in the moment and drinking, it sounds like that's the only time that you said you're feeling happy*»), and the problems associated with C's alcohol use («*So, you're here because your wife feels that you have a problem with drinking*»).

**(23) 9 Family Dynamics and Identity.** K: ['family', 'your family', 'the family', 'your', 'the', 'family and', 'and', 'as', 'you', 'with']. N: 102. T discusses C's relationships within their family («*So just tell me a little bit about your family*», «*I would like you to frame the connection between you and your family*»), including in the context of C's identity («*I mean we started from the topic of identity and where do I belong or where do I stay, so you made some comments on that and then we came to the family topic that, as far as I remember*»).

**(24) 6 Therapy and Father Relationships.** K: ['therapy', 'dad', 'therapist', 'your dad', 'counseling', 'father', 'your', 'your father', 'to', 'you']. N: 108. T reflects on psychotherapy («*I had one of my professors tell me he wants therapy to be unique in that it's two people focusing on one person*») and addresses C's relationship with their father («*Were you close with your dad?*»).

**(25) 15 Relationship with Your Mom.** K: ['mom', 'your mom', 'your', 'mother', 'with your', 'your mother', 'her', 'mom and', 'with', 'you']. N: 81. T explores C's relationship with their mother («*What effect would that have on your relationship with your mom?*»), noting both the differences («*That sounds like you and your mom were different*») and similarities («*Yeah, and you got that from your mom*») between C and their mother.

**(26) 11 Expressions of Happiness and Joy.** K: ['happy', 'like', 'glad', 'joy', 'fun', 'am', 'like that', 'really', 'to hear', 'am so']. N: 85. T encourages C by acknowledging their positive attitude («*Yeah, you seem like a really happy person right now*»), celebrates C's progress in therapy («*I'm really glad to hear that and it looks like you're sleeping better too*»), and reflects on the topic of happiness and joy («*You need to do things that bring you joy to be happy, but you need to be happy to be able to do that*»).

**(27) 20 Understanding and Managing Pain.** K: ['pain', 'painful', 'symptoms', 'the pain', 'the', 'physical symptoms', 'hurts', 'it', 'hurt', 'physical']. N: 70. T discusses overcoming pain with C, primarily physical pain («*You started feeling physical pain*», «*Seems like you're more or less making things worse in an effort to avoid the pain that you're going to suffer*»).

**(28) 13 School Transitions and Experiences.** K: ['school', 'third', 'college', 'the school', 'class', 'the', 'teacher', 'school you', 'grade', 'in']. N: 83. T discusses C's educational experience, such as their forms of involvement in school activities («*Did you do those activities in high school?*») and their experiences attending different types of schools («*How was that transition for you going from public school to private school?*»).

**(29) 26 Shadows of Reflexive Experience.** K: ['it', 'it is', 'like', 'is', 'shadow', 'that', 'of', 'weird', 'kind of', 'kind']. N: 48. T uses various forms of metaphor in their interpretations («*So, it sounds like that part in the shadows is recognizing that this system*», «*It's something like a mist, like a mood that influences concrete things that come up in your mind*», «*It's kind of like sitting on a train watching the scenery go by and you're watching it*»).

**(30) 24 Safety and Self-Protection Insights.** K: ['safe', 'safety', 'you', 'of', 'protectors', 'protect', 'to', 'the', 'that', 'protecting']. N: 49. T discusses various aspects of C's self-protection («*What can you do that's going to give yourself a feeling of safety?*», «*And sometimes that ties in with, you know, there might be places that you can go to get safe*»).

**(31) 10 Self Talk and Self Care.** K: ['yourself', 'self', 'to yourself', 'self talk', 'you', 'to', 'would', 'myself', 'remind', 'about yourself']. N: 90. T explores various aspects of C's self-talk and self-care («*What kind of things would you say to yourself?*», «*Would you feel comfortable giving that advice to yourself?*»), while also expressing their acceptance of the client («*I'm respecting and owning your current reality and accepting yourself in that rather than trying to persuade yourself that you have to be something other*»).

**(32) 8 Exploring Communication Tools and Techniques.** K: ['say', 'you say', 'you', 'tools', 'what', 'to say', 'system', 'experiment', 'say that', 'it']. N: 102. T highlights C's availability of various communication tools applicable to specific situations («*And you got loads of different tools and techniques to help you do that*»).

**(33) 3 Collaborative Problem Solving Approach.** K: ['we', 'we can', 'we are', 'we have', 'to', 'we will', 'let', 'will', 'and', 'can']. N: 174. T encourages C to engage in collaborative work and establishes the agenda for the therapy sessions («*Do you know what we've been working on just to kind of make sure we're on the same page?*», «*So that's the kind of thing that the conflict you just described is the kind of thing that this works really well for and so that's what we could work on if you want*»).

**(34) 0 Others.** K: ['she', 'not', 'okay', 'that', 'is', 'you', 'to', 'yeah', 'he', 'her']. N: 4117. As before, this category contains various topics that were removed during the work process.

**(35) 2 Job Search and Work Challenges.** K: ['job', 'work', 'you', 'income', 'that', 'working', 'to', 'your job', 'jobs', 'the job']. N: 179. T discusses various aspects related to C's job search («*And let's go back to the job application and what you want to write down about that*») or their current employment («*Okay, I see you adjusted to the new income level with the full-time and part-time jobs together*»).

**(36) 12 Breathing and Meditation Practices.** K: ['breathing', 'meditation', 'breath', 'you', 'yoga', 'to', 'practice', 'do', 'your', 'of']. N: 85. T introduces breathing techniques and meditation («*You notice more about your breathing when you are tuned to your breathing*», «*Meditation where you focus on just paying attention to your breath and what happens to all of that crazy thinking when you're just with your breath*»), including in the context of helping C overcome anxiety («*When you notice that anxiety is starting to heighten and then with your breathing you could do some of those other techniques that we talked about*»).

**(37) 5 Discovering Your True Desires.** K: ['do', 'want', 'you', 'you want', 'to do', 'need', 'do that', 'do you', 'that', 'what']. N: 112. T addresses the topic of C's self-determination («*I mean there is a time when one wants to put one's energy into work and school and you know to figure out who we want to be in the world and what you want to do with your life*»), including in the context of clarifying the goals of the therapy sessions («*What is it that you need and we want to bring that into your life?*»).

**(38) 7 Navigating Complex Friendships.** K: ['relationship', 'friend', 'friends', 'relationships', 'you', 'in', 'with', 'and', 'have', 'that']. N: 106. T addresses the topic of friendship in C's life («*So, you said that you have a lot of friends who you've been talking to and meeting new people*», «*And then there's the relationship issue there*»).

**(39) 1 Emotional Feelings and Connections.** K: ['feel', 'feeling', 'you feel', 'emotion', 'you', 'how', 'emotional', 'emotions', 'sadness', 'and']. N: 264. T is interested in C's current («*Do you feel better?*») or past state («*How did you feel?*»), their attitude toward something in a situation («*How do you feel toward it as you noticed?*»), as well as C's attitude toward T («*And how do you feel towards me?*»).

**(40) 14 Stages of Healing Recovery.** K: ['recovery', 'healing', 'of', 'the', 'process', 'you', 'healing state', 'your', 'your recovery', 'state']. N: 82. T notes that therapy is a gradual process and that there are different stages of recovery («*And yeah, it sounds like to me you are at a different stage of recovery than where you were before*», «*And that's why I try to kind of push your way to recovery*»).

**(41) 27 Understanding Progress and Recovery.** K: ['progress', 'progress that', 'we', 'the', 'the progress', 'that', 'of', 'is', 'it is', 'what']. N: 47. T discusses the importance of progress in therapy («*The next thing from my point of view is when there's a little bit of progress*», «*But I think that's where I'm having real evidence of progress becomes important because if you're trying to kind of persuade yourself that you're better than you were*»).

**(42) 4 Weekly Planning and Meetings.** K: ['week', 'time', 'minutes', 'next week', 'this week', 'next', 'morning', 'will', 'we', 'the']. N: 167. T plans the session with C for next week («*Same time next week*») and also assigns some homework to C («*So, consider those two things and we'll pick up with this next week*»).

**(43) 25 Understanding Chronic Depression Dynamics.** K: ['depression', 'depressed', 'depression and', 'right', 'get', 'the', 'of depression', 'that', 'and', 'when']. N: 48. T addresses the topic of depression and its causes («*We're going to work on kind of getting to know about this depression and how it affects you and maybe the origins of it*»), as well as various methods for dealing with it («*As far as looking at the medications versus looking at the origins of the depression if maybe getting specific about when your last episode was to be able to see if we can get underneath, but we can take either path*»).

**(44) 21 Understanding Anxiety Triggers.** K: ['anxiety', 'of anxiety', 'when', 'when you', 'that', 'of', 'the', 'panic', 'you', 'up']: N: 64. T addresses anxiety, its causes, and C's triggers that may provoke it («*So, when you notice that anxiety comes up for you, where do you normally feel it?*», «*Just within the week when you notice that the anxiety came up because of that control reason*»).

**(45) 28 Navigating Fear and Concerns.** K: ['fear', 'the fear', 'afraid', 'afraid of', 'worried', 'concern', 'about', 'the', 'worry', 'little concerned']. N: 46. T discusses the topic of fear with C («*What are you afraid of?*») and its dangers to mental health («*Can I tell you more about the danger of fear?*»), as well as works with C's fear using various techniques («*Turn your attention to the Fear Part in your chest*», «*Yeah, let the fear know that you are not going to forget about it*»).

**(46) 39 Self-Criticism and Social Dynamics.** K: ['they', 'themselves', 'people', 'of', 'they are', 'or', 'that', 'kind of', 'kind', 'that they']. N: 29. T explores the aspects of social dynamics in C's life («*But as people do, they have a way to kind of signal that they're not into that*») and their dependence on the opinions of others («*Do you have any evidence to do something that makes you think that they'll be critical of you?*»).

**(47) 46 Effective Communication Strategies.** K: ['communication', 'they', 'the one', 'them', 'signals', 'how to', 'how', 'sending', 'to express', 'communicate']. N: 20. T discusses with C issues related to building effective communication strategies («*There is a pathway to communication*», «*Don't forget also to have you send out signals?*»).

### 5.3 Proximity of Topics Between Classical and Modern Therapists

The proximity of topics between classical and modern therapists was determined based on cosine similarity. This method calculates the cosine of the angle between two vectors in a multi-dimensional space, measuring how similar the topics are in terms of their content and context. By applying this technique, we can quantify the degree of similarity between the thematic representations of classical and modern therapists, allowing for a comparative analysis of their speech patterns and therapeutic approaches.

Cosine similarity helps demonstrate the most prominent topics in the speech of the two groups of therapists, which likely reflect the core issues clients bring to therapy. For brevity, we will highlight only those topics with a coherence level ranging from 0.9 to 1.0. These include:

- Understanding and combating fear: exploring its nature and addressing anxiety-inducing issues: (24)6 and (45)28;
- Understanding, expressing, and managing anger: (36)37 & (12)29;
- Work-related fear, anxiety, and challenges: (11)33 & (35)2;
- Therapeutic session planning: (42)4 & (43)1;
- Navigating relationship and friendship complexities: (2)15 & (38)7;
- Educational concerns: (15)16 & (28)13;
- Relationships with mothers: (1)21 & (25)15;
- Exploring and overcoming resentment, bitterness, and pain: (8)12 & (27)20;
- Aging, childhood reflections, and generational perspectives: (28)26 & (15)23;
- Exploring possibilities, potential outcomes, and uncertainties: (35)30 & (1)32;
- Self-acceptance, confidence, and the impact of self-criticism on social dynamics: (18)5 & (46)39;
- Understanding personal goals, satisfaction, and the process of goal-setting and achievement: (27)32 & (21)40.

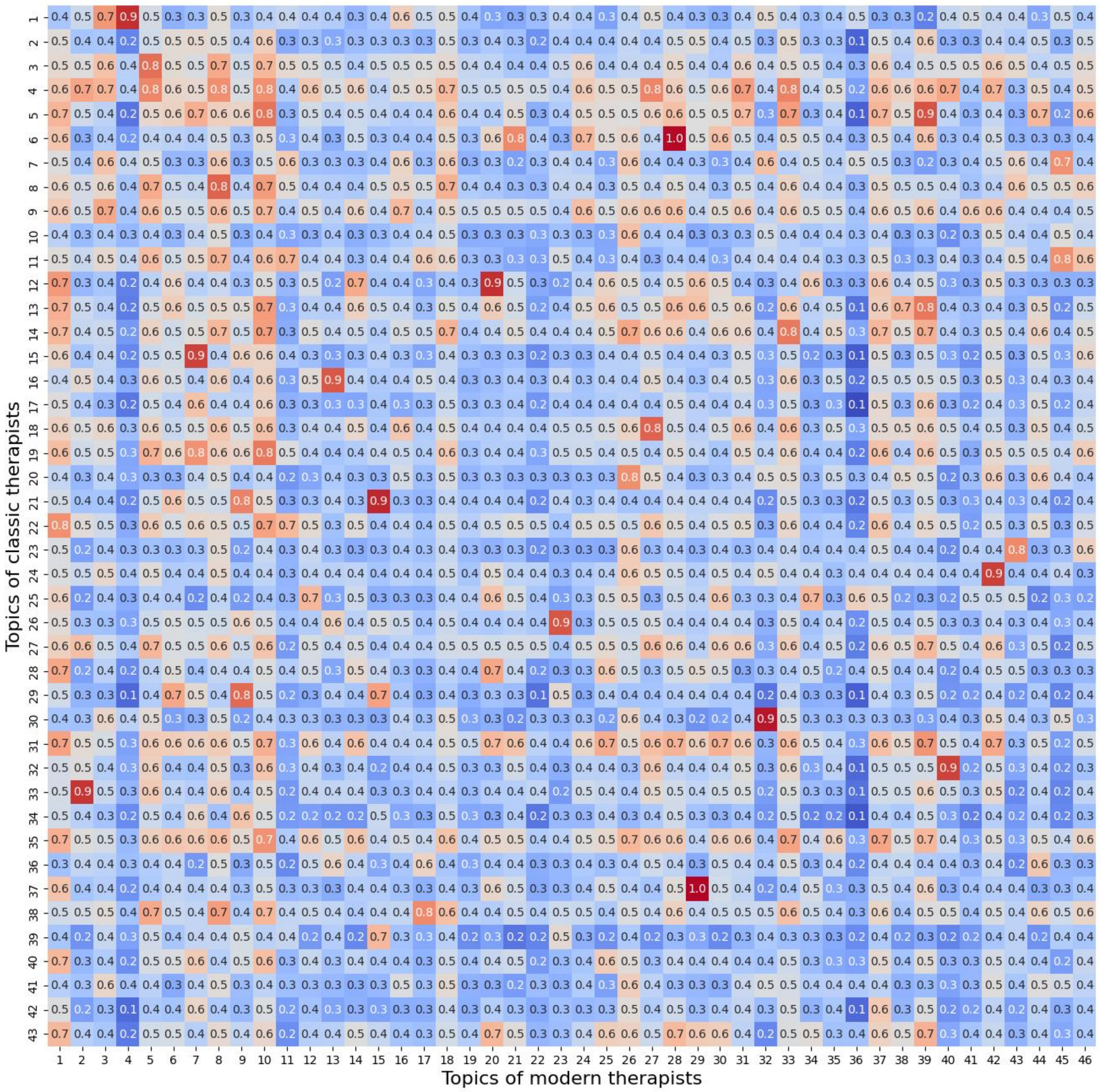

**Fig. 4** A heatmap of cosine similarity between topics of classical and modern therapists

## 6. Conclusion

In conclusion, the ability of BERTopic to perform unsupervised learning makes it a powerful tool for psychotherapy content analysis across a wide range of contexts. As a topic modeling technique based on BERT (LLM-based) embeddings, it offers an effective approach to uncovering meaningful and interpretable themes from large volumes of unstructured text. This capability is especially valuable in psychotherapy, where therapists' language is often complex, nuanced, and highly individualized. By applying BERTopic to therapist remarks and topic representation optimization by LLM specializing in NLP, such as GPT, offers a novel approach to extract distinct topics, enabling deeper insights into common therapeutic practices, as well as emerging trends in therapy sessions.

The results of this study demonstrate the potential of BERTopic to highlight the varied approaches used by therapists, whether classical or modern, and provide a structured analysis of how verbal interventions influence the therapeutic process. This method enhances our understanding of the ways therapists guide sessions, manage emotions, and address client concerns, offering a more systematic, data-driven approach to studying psychotherapy. Ultimately, BERTopic opens up new opportunities for both clinical practice and therapist training, offering a more efficient and unbiased alternative to traditional manual coding methods. Future research can further explore its potential to refine and enhance psychotherapy practices, contributing to the growing body of knowledge on applying machine learning techniques in mental health contexts.

## 7. Future Work

A promising next step in this line of research would be the exploration of client topics in addition to the therapist's speech. Analyzing client remarks could provide valuable insights into the most frequent concerns and requests that clients bring to therapy. This would, in turn, help identify key areas of knowledge and focus for training future therapists, allowing for a more targeted and informed approach to clinical education. By understanding common client needs and expressions, therapists could be better equipped to address specific concerns in a more personalized and effective manner.

Another potential avenue for future work is studying the dynamic development of topics throughout a single therapeutic session. Examining how topics evolve over time could offer a deeper understanding of the therapeutic process and the impact of interventions at different stages of a session. Tracking these changes would provide real-time insights into how therapeutic approaches are working and where adjustments might be necessary to enhance client outcomes.

Ultimately, the automated identification of session topics could pave the way for the development of digital assistants for therapists. Such tools could analyze and record the most significant moments of a therapy session, offering therapists suggestions for further exploration and intervention. By integrating this technology into clinical practice, therapists would be able to receive immediate feedback and recommendations, optimizing the therapeutic process and improving the quality of care provided to clients. This innovation has the potential to significantly enhance both the efficiency and effectiveness of psychotherapy in the future.